\documentclass[nohyperref]{article}

\usepackage{microtype}
\usepackage{graphicx}
\usepackage{booktabs} 
\usepackage{caption}
\usepackage{subcaption}

\usepackage{hyperref}

\usepackage[accepted]{icml2022}

\usepackage{amsmath}
\usepackage{amssymb}
\usepackage{mathtools}
\usepackage{amsthm}

\usepackage[capitalize,noabbrev]{cleveref}

\theoremstyle{plain}
\newtheorem{theorem}{Theorem}[section]
\newtheorem{proposition}[theorem]{Proposition}

\theoremstyle{definition}

\theoremstyle{remark}

\usepackage[textsize=tiny]{todonotes}

\newif\ifcomments
\commentstrue
\ifcomments\newcommand{\comments}[1]{#1}\else\newcommand{\comments}[1]{}\fi

\definecolor{clrgp}{rgb}{.9,0,.9}
\definecolor{red}{rgb}{.8,0,0}
\definecolor{blue}{rgb}{0,0, 0.8}

\icmltitlerunning{Variational Nearest Neighbor Gaussian Process}

\DeclareMathOperator*{\expectedvalue}{\mathbb{E}}

\DeclareMathOperator*{\kltext}{KL}

\newcommand{\bigo}[1]{\ensuremath{\mathchoice{\mathcal O \! \left( #1 \right)}}{\mathcal O(#1)}{\mathcal O(#1)}{\mathcal O(#1)}}

\newcommand{\Evover}[2]{\ensuremath{\expectedvalue_{#1} \left[ #2 \right]}}

\newcommand{\kl}[2]{\ensuremath{\kltext \left[ #1 \: \Vert \: #2 \right]}}

\newif\ifboldmatrix
\boldmatrixtrue

\ifboldmatrix\newcommand{\boldmatrix}[1]{\mathbf{#1}}\else\newcommand{\boldmatrix}[1]{#1}\fi
\ifboldmatrix\else\fi
\ifboldmatrix\else\fi

\newcommand{\bb}{\ensuremath{\mathbf{b}}}

\newcommand{\bfn}{\ensuremath{\mathbf{f}}}
\newcommand{\bk}{\ensuremath{\mathbf{k}}}

\newcommand{\bmm}{\ensuremath{\mathbf{m}}}

\newcommand{\bs}{\ensuremath{\mathbf{s}}}

\newcommand{\bu}{\ensuremath{\mathbf{u}}}

\newcommand{\bx}{\ensuremath{\mathbf{x}}}
\newcommand{\by}{\ensuremath{\mathbf{y}}}
\newcommand{\bz}{\ensuremath{\mathbf{z}}}

\newcommand{\bK}{\ensuremath{\boldmatrix{K}}}
\newcommand{\bL}{\ensuremath{\boldmatrix{L}}}

\newcommand{\bS}{\ensuremath{\boldmatrix{S}}}

\begin{document}

\twocolumn[
\icmltitle{Variational Nearest Neighbor Gaussian Process}




\begin{icmlauthorlist}
\icmlauthor{Luhuan Wu}{stats}
\icmlauthor{Geoff Pleiss}{zuckerman}
\icmlauthor{John Cunningham}{stats,zuckerman}
\end{icmlauthorlist}

\icmlaffiliation{stats}{Department of Statistics, Columbia University}
\icmlaffiliation{zuckerman}{Zuckerman Institute, Columbia University}

\icmlcorrespondingauthor{Luhuan Wu}{lw2827@columbia.edu}

\icmlkeywords{Machine Learning, ICML}

\vskip 0.3in
]



\printAffiliationsAndNotice{}  

\begin{abstract}
Variational approximations to Gaussian processes (GPs) typically use a small set of inducing points to form a low-rank approximation to the covariance matrix. In this work, we instead exploit a sparse approximation of the precision matrix. We propose variational nearest neighbor Gaussian process (VNNGP), which introduces a prior that only retains correlations within $K$ nearest-neighboring observations, thereby inducing sparse precision structure. Using the variational framework, VNNGP's objective can be factorized over both observations and inducing points, enabling stochastic optimization with a time complexity of $O(K^3)$. Hence, we can arbitrarily scale the inducing point size, even to the point of putting inducing points at every observed location. We compare VNNGP to other scalable GPs through various experiments, and demonstrate that VNNGP (1) can dramatically outperform low-rank methods, and (2) is less prone to overfitting than other nearest neighbor methods.
\end{abstract}

\section{Introduction}
\label{sec:intro}
Gaussian processes (GPs) provide rich priors over functions \citep{williams2006gaussian}.
While the GP posterior can be computed in closed form for regression models,
variational posterior approximations \citep{titsias2009variational,hensman2013gaussian, van2020framework}
have become increasingly popular as they offer numerous advantages.
First,
variational methods can utilize stochastic minibatching \citep{hensman2013gaussian},
enabling inference for large-scale datasets.
Second, variational methods are compatible with non-conjugate likelihoods \citep{hensman2015mcmc} and simplify inference for interdomain observation models \citep{lazaro2009inter,alvarez2010efficient,van2020framework,wu2021hierarchical}.
Finally,
black box variational methods \citep{ranganath2014black} make it possible to use GP as components of larger  models, such as Deep Gaussian Processes \citep{damianou2013deep}.

To apply variational methods to large-scale GP,
it is often necessary to make additional approximations \citep{titsias2009variational, hensman2017variational}.
This is because making GP inference with $N$ observations incurs a $\bigo{N^3}$ computation.
A popular variational approach that aids this issue is the Stochastic Variational Gaussian Process method (SVGP) \citep{hensman2013gaussian},
which essentially forms a low-rank approximation to the prior covariance.
These approximations offer well understood predictive properties \citep{bauer2017understanding}, probable performance guarantees \citep{burt2019rates}, and competitive performance on numerous datasets.
However, there are also many instances when low-rank approximations are ill suited.
For example, spatiotemporal datasets often have \emph{intrinsically low lengthscale},
meaning that the data can vary rapidly with small changes in space and time.
In these settings, low-rank approximations cannot capture fast variations,
often resulting in low fidelity or ``blurry'' predictions \citep[e.g.][] {datta2016hierarchical,pleiss2020fast,wu2021hierarchical}.

In this paper, we are interested in making variational GP methods more amenable to spatiotemporal data and other domains with  intrinsically low lengthscale, while retaining (and even improving) scalability.
To do so, we replace SVGP's low-rank approximation of the prior covariance matrix with a 
\emph{sparse approximation of the prior precision matrix}.
Whereas a low-rank approximation assumes that observations are explained through a small number of global latent variables,
a sparse precision approximation instead assumes that observations are conditionally independent given their nearest neighboring observations.  
Mathematically, the joint GP prior is approximated by the product of one-dimensional conditionals, each of which only depends on a small subset of preceding observations based on some predetermined ordering.
This assumption is quite reasonable in the context of spatiotemporal modeling.
For example, it is reasonable to assume that the weather in Los Angeles and Maine are conditionally independent, given neighboring cities in California. 

Sparse precision approximations are common in spatiotemporal modeling  throughout the geostatisics literature \citep[e.g.][]{vecchia1988estimation,stein2004approximating,datta2016hierarchical,guinness2018permutation,finley2019efficient,katzfuss2021general},
where they often achieve higher predictive fidelity than low-rank approximations.
However, most existing approaches target exact or MCMC inference and are not compatible with variational methods.
As a result, it is not straightforward to use sparse precision methods with non-conjugate or interdomain likelihoods, or as part of larger probabilistic models without resorting to sampling.
Recently, \citet{tran2021sparse} made a first key step in this direction. 
They introduce Sparse Within Sparse Gaussian Processes (SWSGP), which combines a similar nearest neighbor approximation with SVGP. 
Instead of exploiting the sparse precision structure, SWSGP introduces sparsity by constructing a hierarchical prior. While successful, we will show an important technical opportunity remains, which we offer here.  

In this work, we propose a novel sparse precision approximation that is compatible with stochastic variational inference.
Similar to SVGP, our proposed method---the \emph{Variational Nearest Neighbor Gaussian Process} ({\bf VNNGP})
\footnotemark
-- formulates the variational objective and predictive distribution through a set of $M$ \emph{inducing points}.  
Unlike SVGP, our method applies 1) a sparse precision approximation to the inducing point distribution and 2) a sparse dependency approximation between observations and inducing points. 
These approximations enable the VNNGP objective to factorize over data \emph{and} inducing points. Consequently, we can minibatch over data and inducing points when evaluating the variational objective, resulting in \emph{constant time computational complexity with respect to both $N$ and $M$}.
This significant reduction makes it possible to scale $M$ well beyond the limits of SVGP, even to the point of placing inducing points at every observed location. Moreover, unlike SWSGP and other sparse approximations \citep{vecchia1988estimation, katzfuss2021general}, the VNNGP prior and variational distribution both constitute valid GPs, enabling it  to more faithfully retain the features of the true GP prior. 
We compare VNNGP to SVGP, SWSGP, and other scalable methods on numerous toy experiments and benchmark datasets.
Our results demonstrate that VNNGP can offer higher fidelity predictions than SVGP and is less prone to overfitting than SWSGP.
These advantages are most pronounced on spatiotemporal datasets (including those with non-conjugate observational models), but hold for high dimensional datasets as well.

\section{Background}
\label{sec:background}

\subsection{Gaussian Processes}
Consider any finite dimensional realization $\bfn = \{f_i\}_{i=1}^N$ of

\footnotetext{VNNGP is implemented in the GPyTorch library. See the example in 
\url{https://docs.gpytorch.ai/en/stable/examples/04_Variational_and_Approximate_GPs/VNNGP.html}}
a Gaussian process at locations $\bx_1, \bx_2,\cdots , \bx_N \in \mathbb{R}^D$: 
\begin{align}
\label{eq:gp-latent}
\begin{split}
f(\cdot) &\sim \mathcal{GP} (0, k(\cdot, \cdot)) \\
 f_i&  \equiv f(\bx_i) \qquad \textrm{ for } i=1:N 
\end{split}
\end{align}
where $k(\cdot, \cdot)$ is a kernel function that encodes the properties of the prior. We can always factorize the joint probability of a GP realization into a chain of conditional probabilities subject to some ordering, for example
\begin{align}
\label{eq:gp-conditionals}
    p(\bfn) &= p(f_1) \prod_{i=2}^N p(f_i | \bfn_{1:i-1}), 
\end{align}
where $p(f_1) = \mathcal{N}( f_1 | 0, k_{1,1})$, 
\begin{align}
\label{eq:gp-predictive}
\begin{split}
    p(f_i | \bfn_{1:i-1}) &= \mathcal{N} ( f_i | \bk^\top_{1:i-1,i} \bK_{1:i-1, 1:i-1}^{-1}  \bfn_{1:i-1},  \\ 
    & \qquad k_{i,i} - \bk^\top_{1:i-1,i} \bK_{1:i-1, 1:i-1}^{-1} \bk_{1:i-1,i})
\end{split}
\end{align}
and we define $k_{i,i} \equiv k(\bx_i, \bx_i)$, $\bk_{1:i-1, i} \equiv k(\bx_{1:i-1}, 
\bx_i)$, and $\bK_{1:i-1, 1:i-1} \equiv k(\bx_{1:i-1,i}, \bx_{1:i-1})$. The following notations will apply this convention accordingly.

\subsection{Sparse Precision Approximations}
GP approximations based on sparse precision matrices remain popular for years, largely in the geostatistics community \citep{vecchia1988estimation, datta2016hierarchical, finley2019efficient, katzfuss2021general}. They consider the following approximation to Eq~\ref{eq:gp-conditionals} 
\begin{align}
\label{eq:sparse-gp-conditionals}
        p(\bfn ) &\approx p(f_1) \prod_{i=2}^N p(f_i | \bfn_{n(i)}),  
\end{align}
where $p(f_1) = \mathcal{N} (f_1 | 0, k_{1,1})$, 
\begin{align}
\begin{split}
    p(f_i | \bfn_{n(i)}) &= \mathcal{N} (f_i | \bk^\top_{n(i),i} \bK_{n(i), n(i)}^{-1} \bfn_{n(i)}, \\
    & \qquad k_{i,i} - \bk^\top_{n(i),i} \bK_{n(i), n(i)}^{-1} \bk_{n(i), i})
\end{split}
\end{align}
and $n(i)$ denotes the indices of $K$ nearest neighbors of $\bx_i$ in $\{ \bx_k\}_{k=1}^{i-1}$. Typically $K$ is set to be much smaller than $N$, and when $K=N$ Eq~\ref{eq:sparse-gp-conditionals} recovers the original GP. This approximation  results in a sparse Cholesky factor of the precision matrix \citep{datta2021sparse}. As a consequence, the model complexity scales $O(NK^3)$. It has seen many successful applications in spatiotemporal problems, e.g.  predicting PM polutant levels arocss Europe \citep{datta2016nonseparable}, predicting forest canopy height \citep{finley2017applying}, estimating forest biomass \citep{taylor2019spatial}, etc.  

\subsection{Stochastic Variational Gaussian Processes}
Stochastic variational Gaussian process (SVGP) \citep{hensman2013gaussian} defines a small set of $M$ inducing points $\bu = \{ u_j\}_{j=1}^M$ which are GP latent variables at locations $\{\bz_j\}_{j=1}^M$. It considers the joint latent generative process: 
\begin{align}
\begin{split}
    &p(\bu, \bfn)  = \mathcal{N}( \begin{pmatrix} \bu \\ \bfn \end{pmatrix} | 0, \begin{pmatrix} \bK_{\bz, \bz} & \bK_{\bz, \bx} \\ \bK_{\bx, \bz} & \bK_{\bx, \bx} \end{pmatrix})
\end{split} 
\label{eq:svgp-generative-process}
\end{align}
and an independent observation model 
\begin{align}
& p(\by | \bfn ) = \prod_{i=1}^N p(y_i|f_i)
\label{eq:svgp-likelihood}
\end{align}
where $p(y_i|f_i)$ is an arbitrary likelihood function.  
SVGP is optimized by maximizing the evidence lower bound (ELBO) given below:
\begin{align}
\mathcal{L}_{\textrm{SVGP}} &= 
    \sum_{i=1}^N \Evover{q(f_i)}{ \log p(y_i | f_i) } - \kl{q(\bu)}{p(\bu)},
\label{eq:svgp-elbo}
\end{align}
where $q(\bu)$ is the variational posterior for $\bu$
and $q(f_i) = \int p(f_i | \bu) q(\bu) \mathrm d \bu$ is the variational posterior for $f_i$. The SVGP approximation relies on the low-rank matrix $\bK_{\bx,\bz} \bK^{-1}_{\bz,\bz} \bK_{\bz ,\bx}$ instead of the full-rank matrix $\bK_{\bx, \bx}$ to reduce the size of any matrix inversion to $M$. Therefore, we refer to it as a low-rank approximation.

SVGP has several advantages compared to exact GP and other scalable methods. First, as is shown in Eq~\ref{eq:svgp-elbo}, the ELBO is factorzied over data points. Therefore, it is amenable to stochastic optimization and the computational complexity reduces to $O(M^3)$. Moreover, due to its variational inference nature, SVGP can be applied to problems with  non-conjugate likelihoods, inter-domain observations or latent variable models based on GPs. However, since SVGP assumes $M \ll N$, it may struggle to obtain accurate predictions for large-scale data that are not inherently low-rank structured \citep{wu2021hierarchical, tran2021sparse}. 

\subsection{Other Related Works}
\paragraph{Variational GP inference} was first pursued by \citet{csato2000efficient} and \citet{gibbs2000variational}.
More recent works connect variational inference with scalable methods.
\citet{titsias2009variational} ties inducing point approximations and variational inference,
and \citet{hensman2013gaussian} extends this approach to be compatible with minibatched optimization.
Recent extensions focus on inducing points in linearly transformed domains \citep{lazaro2009inter, van2020framework}, inducing points with exploitable algebraic kernel structures  \citep{wilson2015kernel, wu2021hierarchical},  and inducing points with separable structure \citep{cheng2017variational,salimbeni2018natural,shi2020sparse}.
Beyond inducing point approximations, \citet{hensman2017variational} and \citet{john2018large} combine variational inference with finite basis approximations.

\paragraph{Nearest neighbor approximations.}
In addition to the sparse precision approximations pursued by the geostatistics community (see introduction), the machine learning community has proposed nearest neighbor GP approximations \citep{kim2005analyzing, nguyen2008local, gramacy2015local, park2018patchwork, jankowiak2021scalable}. Besides SWSGP which is introduced before, we review three methods that are most related to our method. 
\citet{bui2014tree} proposes a sparse prior by imposing a tree or chain structure over inducing points and calibrating the posterior approximation with a KL divergence minimization. This approach uses a message-passing inference algorithm and learns the hyperparameters by maximizing the marginal likelihood over inducing points. 
Stochastic gradient descent GP (sgGP) forms a (biased) stochastic GP objective that optimizes $K$-nearest-neighboring data points per iteration \citep{chen2020stochastic}. 
This method is an approximation to exact GP inference, and therefore is limited to Gaussian likelihood.
\citet{liu2019amortized} use amortized variational inference for data points within a small neighborhood, where the variational covariance is constructed to have a sparse Cholesky factor. The latter two methods have computational complexity cubic in the number of nearest neighbors.

\section{Variational Nearest Neighbor GP}
\label{sec:method}
In this section, we develop VNNGP, a highly scalable variational GP method. VNNGP considers the same observation model in Eq~\ref{eq:svgp-likelihood} and makes the following  nearest neighbor approximations to the latent generative process
\begin{align}
\label{eq:vnngp-prior}
    p(\bu)  &\approx \prod_{j=1}^M p(u_i | \bu_{n(j)}), \quad  
    p(\bfn | \bu) \approx \prod_{i=1}^N p (f_i | \bu_{n(i)}) 
\end{align}
where  each conditional denotes a standard GP predictive distribution, $n(j)$ denotes the inducing point indices corresponding to $\bz_j$'s (at most) $K$ nearest neighbors selected from $\{ \bz_1, \cdots, \bz_{j-1}\}$ (with a special case of $n(1) = \emptyset$), 
and we overload the notation $n(i)$ to denote the inducing point indices of $\bx_i$'s $K$ nearest neighbors in all  $\{\bz_j\}_{j=1}^{M}$. 

\begin{figure}[!htp]
    \centering
    \includegraphics[scale=0.25]{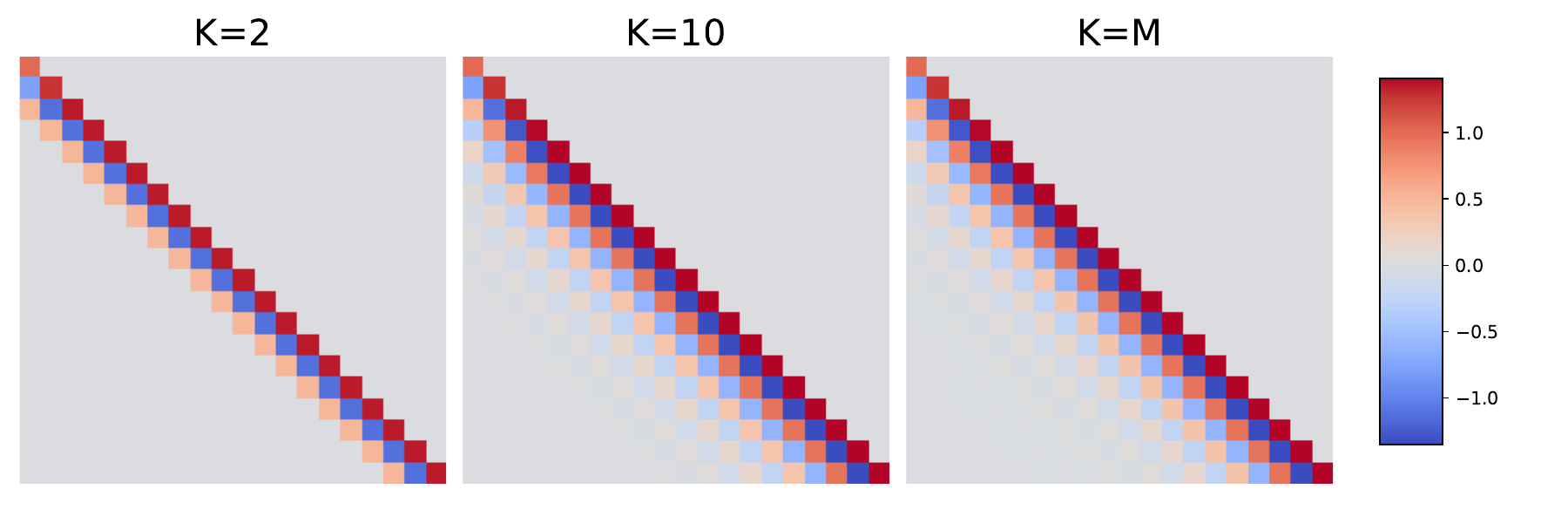}
    \vspace{-1em}
    \caption{The Cholesky factor of prior precision matrix for $M=20$ inducing points (with even spacing of 1) by VNNGP with different number of nearest neighbors $K$. We use a squared exponential kernel with outputscale 1 and lengthscale 1. For each row of the Cholesky factor, at most $K+1$ elements are non-zero, and when $K=10$ the approximation is qualitatively close to the exact one.}
    \vskip -0.1in
    \label{fig:precision-approximation}
\end{figure}

Because of this construction, VNNGP forms a sparse approximation to the prior precision matrix $\bK_{\bz, \bz}^{-1}$ \citep{datta2021sparse}.  Its Cholesky factor has at most $K+1$ non-zero elements per row, corresponding to the $K$-nearest-neighbor dependency structure. 
Figure~\ref{fig:precision-approximation} plots the Cholesky factor of $\bK_{\bz, \bz}^{-1}$ for $M=20$ inducing points with varying number of nearest neighbors $K$. 
We further see that, as $K$ increases, the Cholesky factor becomes denser and approaches the exact one. In this case, $K=10$ yields a quality approximation.    Note that the sparsity pattern of the Cholesky factor depends on the ordering of inducing points and the nearest neighbor sets. In Figure~\ref{fig:precision-approximation} we use the coordinate ordering of 1-dimensional inducing points, thereby rendering a banded sparsity pattern.

\paragraph{Ordering of inducing points.} The nearest neighbor approximation in Eq~\ref{eq:vnngp-prior} is subject to a particular ordering of inducing points. In all following experiments, we apply random ordering and we find that different orderings do not lead substantial differences on VNNGP's performance.  

\paragraph{Variational family.}  We consider a special choice of the variational distribution $q(\bfn, \bu)$ that will greatly simplify the computation without sacrificing much prediction accuracy. Specifically, we use a mean-field variational approximation for $\bu$ and approximate the posterior for $\bfn$ with the nearest neighbor prior: 
\begin{align}
    q(\bu) &= \prod_{j=1}^M q(u_j) = \prod_{j=1}^M \mathcal{N} (u_j | m_j, s_j), \label{eq:qu-mf} \\
    q(\bfn) &= \! \prod_{i=1}^N \! q(f_i) \! = \! \prod_{i=1}^N \! \int \! p(f_i | \bu_{n(i)}) q(\bu_{n(i)}) \mathrm d \bu_{n(i)} \label{eq:qf}. 
\end{align}
The choice of mean-field or sparse variational approximations are common for large-scale GP when full-rank approximations are impractical \cite{liu2019amortized, wu2021hierarchical, tran2021sparse}. Empirically, we found that VNNGP with a mean-field variational distribution outperforms SVGP with a full-rank one for most cases. 
However, we also discuss the extension to a more expressive variational family in appendix~\ref{appendix:sec-variational-family-extension}. 
\paragraph{Optimization objective} is to maximize the VNNGP's ELBO as follows 
\begin{align}
\begin{split}
    \mathcal{L}_{\textrm{VNNGP}} &= \underbrace{  \sum_{i=1}^N \Evover{q(f_i)}{ \log p(y_i | f_i)}}_{\textrm{data likehood}}  - \underbrace{\kl{q(\bu)}{p(\bu)}}_{\textrm{KL divergence}}, 
\end{split}
\label{eq:vnngp-elbo}
\end{align}

Note that the ELBO breaks into a data likelihood term and a KL divergence term. In the following , we first discuss the computation of these two terms. After that, we present the advantages of VNNGP. Finally, we compare VNNGP to other related works based on nearest neighbor approximations.

\subsection{Computing the Data Likelihood Term}
\label{sec:method-data-likelihood}
Similar to SVGP, the data  likelihood term for VNNGP factorizes over data points. Unlike SVGP that considers all $M$ inducing points in making the prediction for one data point, VNNGP makes use of the sparse prior in  Eq~\ref{eq:vnngp-prior} such that the prediction is based on only $K$ nearest inducing points. As a result, VNNGP's  data likelihood term is reduced to 
\begin{align}
\label{eq:data-likelihood-term}
    \sum_{i=1}^N \int p(f_i | \bu_{n(i)}) q(\bu_{n(i)}) \log p(y_i | f_i) d f_i d \bu_{n(i)}
\end{align}
where only $K$ nearest inducing points are involved in computing the expected log likelihood for each data point. 

\subsection{Computing the KL Divergence}
\label{sec:method-kl-divergence}
We now show how we can write KL divergence as a summation over inducing points. Note that it is not possible for the standard SVGP: SVGP's KL requires accessing  parameters for all inducing points at the same time. 

We first break the prior into a chain of conditionals using the nearest neighbor assumption in Eq~\ref{eq:vnngp-prior} and utilize the mean-field assumption in Eq~\ref{eq:qu-mf}: 
\begin{align}
    \kl{q(\bu)}{p(\bu )} 
    &= \kl{\prod_{j=1}^M q(u_j) } {\prod_{j=1}^M p(u_j | \bu_{n(j)})}
    \nonumber \\
    &= \Evover{q(\bu)}{ \sum_{j=1}^M \log\frac{q(u_j)}{p(u_j | \bu_{n(j)})} }
\end{align}
exchanging the order of summation and expectation and integrating out irrelevant variables
\begin{align}
  &= \sum_{j=1}^M \Evover{q(\bu_{n(j)})}{ \mathbb{E}_{q(u_j)}\left[ \log \frac{q(u_j)}{p(u_j | \bu_{n(j)})} \right] } \\
  &= \sum_{j=1}^M \Evover{q(\bu_{n(j)})}{ \kl{q(u_j)}{p(u_j | \bu_{n(j)})} } \label{eq:vnngp-kl-factorization} 
\end{align}
where 
\begin{align}
    p(u_j | \bu_{n(j)}) &= \mathcal{N} (u_j |\bk_{n(j),j}^\top  \bK_{n(j), n(j)}^{-1} \bu_{n(j)}, \\
    & \qquad k_{j,j}-\bk^\top_{n(j),j} \bK_{n(j), n(j)}^{-1} \bk_{n(j),j}). 
\end{align}

\subsection{Advantages}
\label{sec:method-advantages}
Having factorized the data likelihood term and KL divergence over data points and inducing points, we are now ready to present VNNGP's key advantages as follows. 

\paragraph{Stochastic optimization.} The summation structure of VNNGP's ELBO, as detailed in Eq~\ref{eq:data-likelihood-term} and \ref{eq:vnngp-kl-factorization}, makes it immediately available for stochastic optimization: in every training iteration, we randomly sample a mini-batch of training data indices $\mathcal{I} = \{i_k\}_{k=1}^{N_b}$ and a mini-batch of inducing point indices $\mathcal{J} = \{ j_l\}_{l=1}^{M_b}$; we then optimize the unbiased estimate of the ELBO as follows 
\begin{align}
\begin{split}
    &\mathcal{L}_{\textrm{VNNGP}}  \approx 
    \frac{N}{N_b} \sum_{i \in \mathcal{I}} \Evover{q(f_i)}{ \log p(y_i | f_i) }  \\
    & \quad   - \frac{M}{M_b} \sum_{j\in \mathcal{J}} \Evover{q(\bu_{n(j)})}{ \kl{q(u_j)}{p(u_j | \bu_{n(j)})} }.
\end{split}
\label{eq:vnngp-stochastic-objective}
\end{align}
We include the exact computation of ELBO in \cref{appendix:vnngp-elbo-computation}. 
Note that the sampling distributions for data points and inducing points can be arbitrary as long as both remain marginally a uniform distribution. We emphasize that this is a key advantage over SVGP: we are not able to perform stochastic optimization for inducing points with SVGP.

\paragraph{Computational complexity.} 
Once the nearest neighbor  structure is computed, optimizing the objective in Eq~\ref{eq:vnngp-stochastic-objective}  requires $O((N_b + M_b) K^3)$ computations for inverting $K\times K$ kernel matrices induced by nearest neighbor mini-batches. See \cref{appendix:vnngp-elbo-computation} for mathematical details. 

The computational overhead comes from determining  nearest neighbor structures for observations and inducing points. The brute-force complexity of finding any point's $K$ nearest neighbors within $M$ points is $O(KM)$. With the help of modern similarity search packages such as \citet{JDH17}, we are able to dramatically speed up billion-scale similarity search with gpus. For example, for medium-sized datasets, e.g. Protein ($N=25.6$K, $D=9$), it takes no more than 30 seconds to build up nearest neighbor structures with $K=256$ and $M=N$  on an NVIDIA RTX2080 gpu; for the largest dataset we experiment with (Covtype: $N=372$K and $D=54$), it takes approximately 12 minutes with $K=256$ and $M=N$. 

\paragraph{More inducing points.}
Since the training complexity is free of $N$ and $M$, we are able to place  inducing points at every observed location and scale $M$ to $N$. We enjoy several benefits by doing so. First, we greatly boost the model capacity, as opposed to a low-rank approximation where several observations ``share'' one inducing point. Moreover, we avoid optimizing inducing point locations as the standard SVGP training procedure requires. This is a huge advantage since it saves training cost and the optimization of inducing locations is in general non-convex and more difficult. Finally, we only need to compute  the nearest neighbor structure once before training, or otherwise we need to recompute it frequently since inducing locations are being updated.

\subsection{Comparison to Related Methods}
\label{sec:compare-to-related-methods}
The method most related to VNNGP is the Sparse Within Sparse Gaussian Process (SWSGP) method of \citet{tran2021sparse}. 
SWSGP also builds upon SVGP and  further imposes sparsity over $M$ inducing points $\bu$ by defining a hierarchical prior.
Essentially, in each training iteration, SWSGP randomly samples a minibatch of data point indices $\mathcal{I} = \{i_k\}_{k=1}^{N_b}$ and optimizes the following ELBO
\begin{align}
\begin{split}
   & \mathcal{L}_{\textrm{SWSGP}}  = \frac{N}{N_b} \sum_{i \in \mathcal{I}}  \Evover{q(f_i)}{\log p(y_i | f_i)}  \\
    &\qquad - \frac{1}{N_b}\sum_{i \in \mathcal{I}} \kl{q(\bu_{n(i)})}{p(\bu_{n(i)})}), 
\end{split}
\label{eqn:swsgp_objective}
\end{align}
where $n(i)$ are indices of inducing point locations from $\{\bz_j\}_{j=1}^M$ that are top-$K$ nearest to $\bx_i$. 
Similar to VNNGP, evaluating Eq.~\ref{eqn:swsgp_objective} requires $O(N_bK^3)$ computation. 

There are two key differences between VNNGP and SWSGP. First, the two methods introduce sparsity through a different generative process. SWSGP applies a hiearchical prior; Consequently, the marginal prior for either $p(\bfn)$ or $p(\bu)$ is no longer Gaussian. VNNGP, on the other hand, considers a sparse approximation to the prior precision matrix. The resulting process is still a GP but under an approximated kernel.
While it is not immediately clear how these differing marginal distribution impact downstream performance, in this sense the VNNGP generative process more faithfully replicates the exact GP prior.

The second difference is the training objective. As can be seen, the first term of Eq.~\ref{eqn:swsgp_objective} (data likelihood) is identical to that of VNNGP (Eq.~\ref{eq:vnngp-stochastic-objective}). However, the KL divergence term is where two objectives differ. SWSGP optimizes a ``local" KL divergence that only involves inducing points within a local neighborhood around current batch of training data.
In fact, we prove (see appendix~\ref{appendix:swsgp-kl-divergence}) that SWSGP's KL divergence always underestimates the SVGP KL divergence: 
\begin{align}
    \kl{q(\bu_{n(i)})}{p(\bu_{n(i)})}) &\leq \kl{q(\bu)}{p(\bu)},  
\end{align} 
for any subset $n(i)$. Additionally, the amount of underestimation depends on size of $n(i)$ . 

We hypothesize that SWSGP's KL, due to its ``local'' characteristic,  may not sufficiently regularize the variational distribution---especially when $K \ll M$---which may make the model more prone to overfitting.
Conversely, though VNNGP's KL does not equal the exact KL, it does (in expectation) consider the joint distribution over all inducing points.  
Optimizing this term will regularize the variational distribution towards the GP prior at all input locations. We investigate our hypotheses in the experiments section.

Another method that also combines nearest neighbor approximations with variational inference is Amortized Inference GP  \citep{liu2019amortized}.
However, their method avoids computing certain terms of the KL divergence, and in doing so makes it impossible to simulatenously optimize kernel hyperparameters using the ELBO.
By contrast, the VNNGP objective enables gradient-based kernel learning.

\begin{figure}[!t]
    \centering
    \includegraphics[scale=0.35]{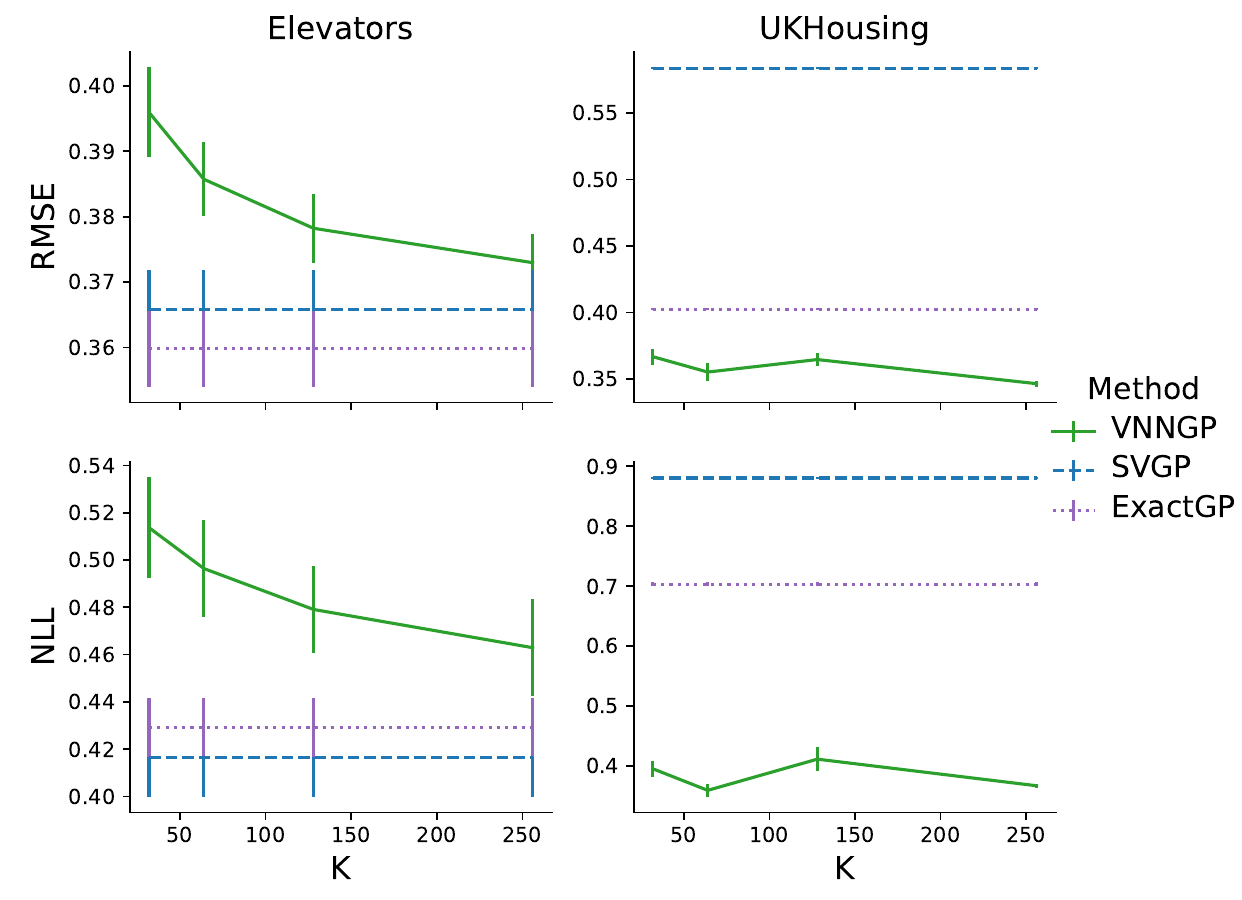}
    \vspace{-1em}
    \caption{Test set RMSE and NLL as a function of number of nearest neighbors $K$ for VNNGP (green) on Elevators (left column) and UKHousing datasets (right column). Results for SVGP (blue dashed) and exact GP (purple dotted) are included as baselines.}
    \label{fig:hyper-sensitivity}
    \vskip -0.1in
\end{figure}

\begin{figure*}[!tp]
\vskip 0.1in
\begin{center}
\includegraphics[scale=0.4]{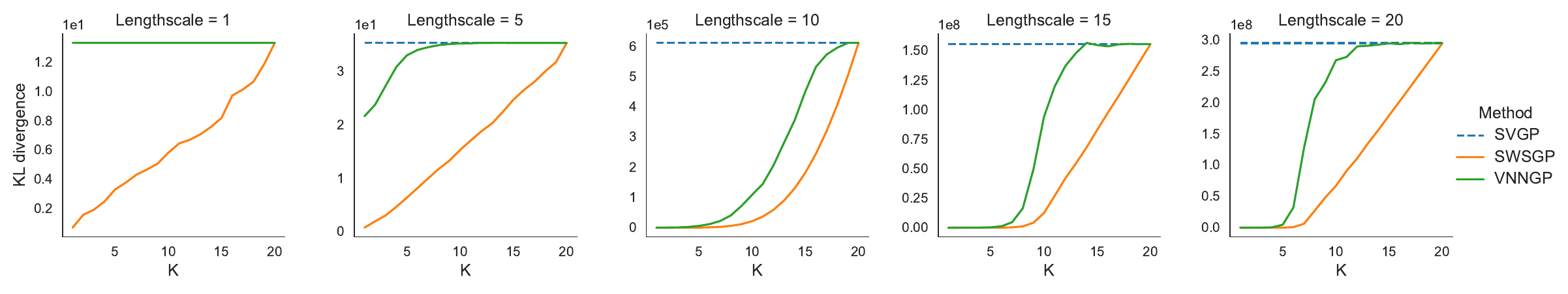}
\vspace{-1em}
\caption{KL divergence computed by VNNGP (green) and SWSGP (orange) as a function of $K$ for different prior kernel lengthscales. Both methods tend to underestimate the KL (SWSGP provably), and converges to the exact value computed by SVGP (blue dotted) as $K \rightarrow M$. Morever, VNNGP's KL is closer to the exact one compared to SWSGP for the same $K$ across all cases.}
\label{fig:kl}
\end{center}
\vskip -0.2in
\end{figure*}

\section{Experiments}
\label{sec:experiment}
We first demonstrate how sensitive our model is to the number of nearest neighbors $K$. 
Recognizing that our method is similar to SWSGP, our next task is to characterize the differences between two methods and investigate our hypotheses in Sec~\ref{sec:compare-to-related-methods}. 
Finally, we evaluate VNNGP and other methods 
on a wide range of real regression and classification datasets. Unless stated otherwise, for all experiments, VNNGP applies a random ordering of inducing points,  VNNGP and SWSGP set inducing points at observed locations and use a mean-field variational approximation, and SVGP uses $1024$ inducing points and a full-rank variational approximation. 

\subsection{Sensitivity to number of nearest neighbors $K$}
\label{sec:exp-hyperparameter-sensitivity}
Our first goal is to see how the number of nearest neighbors $K$ impact VNNGP's performance. We choose two datasets, Elevators with dimension $D=16$ \citep{asuncion2007uci}, and UKHousing (https://landregistry.data.gov.uk/), a spatial dataset with $D=2$, as we expect low-rank approximations to work better in high dimensional settings while nearest neighbor methods are more suitable for spatial problems that are sparse in nature. We fit VNNGP with $K \in [32,64,128,256]$.
Since these two datasets are modeled with Gaussian likelihood, we are able to fit exact GP \citep{wang2019exact} as a baseline. We also include SVGP as a low-rank variational alternative. We present their test set predictive performance,  the root mean squared error (RMSE) and negative log likelihood (NLL), in Figure~\ref{fig:hyper-sensitivity}. 
From the figure, we see that as $K$ increases, both test RMSE and NLL for VNNGP decrease, suggesting a better performance.  By comparing the three methods, we note that exact GP does not always yield the best RMSE or NLL. This result could be due to model misspecification, as observed in previous literature \citep{wang2019exact, potapczynski2021bias}. Moreover, SVGP performs better on the high-dimensional Elevators dataset, 
whereas VNNGP with $K=32$ already outperforms SVGP and is less sensitive to $K$ on the spatial UKHousing dataset.

\subsection{Comparison to SWSGP}
\label{sec:exp-comparison-to-swsgp}
Building on the analysis in \cref{sec:compare-to-related-methods},
we now aim to compare VNNGP to the related SWSGP method.
Recall that the primary difference between these methods is the training objective (the KL divergence term in particular).
We first empirically demonstrate how these two objectives differ, and then study its surprisingly significant impact on model fitting and selection.

\paragraph{KL divergence comparison.}
We first investigate the difference between the VNNGP and SWSGP KL divergence terms.
Specifically, we are interested in how their KL terms change as we vary $K$. 
We initialize VNNGP and SWSGP with the same configuration: the inducing points are placed on a 1-dimensional grid with even spacing of 1, the variational means $\{m_i\}$ are obtained by sampling a GP under a squared exponential kernel with lenghthscale 5 and outputscale 1, and the variational variances $\{s_i\}$ are set to 1.  
For both methods, we compute their KL terms by varying the kernel lengthscale parameter and \# of nearest neighbors $K$. A special note is that when $K=M$, KL by either method should be equal to the KL by SVGP in Eq.~\ref{eq:svgp-elbo}. 

We plot the SWSGP and VNNGP KL terms as a function of $K$ under different kernel lengthscales in \cref{fig:kl}.
From this figure, we first confirm that VNNGP and SWSGP recover the SVGP KL divergence (dotted blue line) at $K=M$. Morever, both methods tend to underestimate KL (recall from \cref{sec:compare-to-related-methods} that SWSGP provably underestimates this term). 
Notably, VNNGP converges rapidly to the true KL divergence, especially for small lengthscale settings. For example, when the lengthscale is 5, the VNNGP KL divergence is indistinguishable from the true KL divergence for all $K\geq 5$. 
We include results under different settings in appendix~\ref{appendix:exp-kl-divergence} and observe similar patterns. 

\paragraph{Fitting models in high observational noise setting.}
We now study the impact of KL divergence on the inferred posterior distribution.
Recall that the KL term of the ELBO regularizes the variational distribution towards the GP prior.
If the regularization is not sufficient, the ELBO will be dominated by the data likelihood term, which could overfit the predictive distribution to the observed data.

Hence, we conduct a simulation experiment in the presence of high observational noise: We draw $N=50$ observations from a GP under a squared exponential kernel with lengthscale 5 and outputscale 5, and add iid noise from $\mathcal{N} (0, 5)$. There are two dense clusters of observations around $x=0$ and $x=50$, and scarcely scattered observations in between. We fit four models---exact GP, SVGP, VNNGP and SWSGP---to the observed data. We fix all hyperparameters corresponding to the true data generating process, and only optimize variational parameters. For the three variational methods, inducing points are set to the locations of observed data ($M=N$).
All variational models use mean-field variational approximations.
Both VNNGP and SWSGP use $K=5$ nearest neighbors. 

\begin{figure}[!t]
\vskip 0.1in
\begin{center}
\includegraphics[scale=0.32]{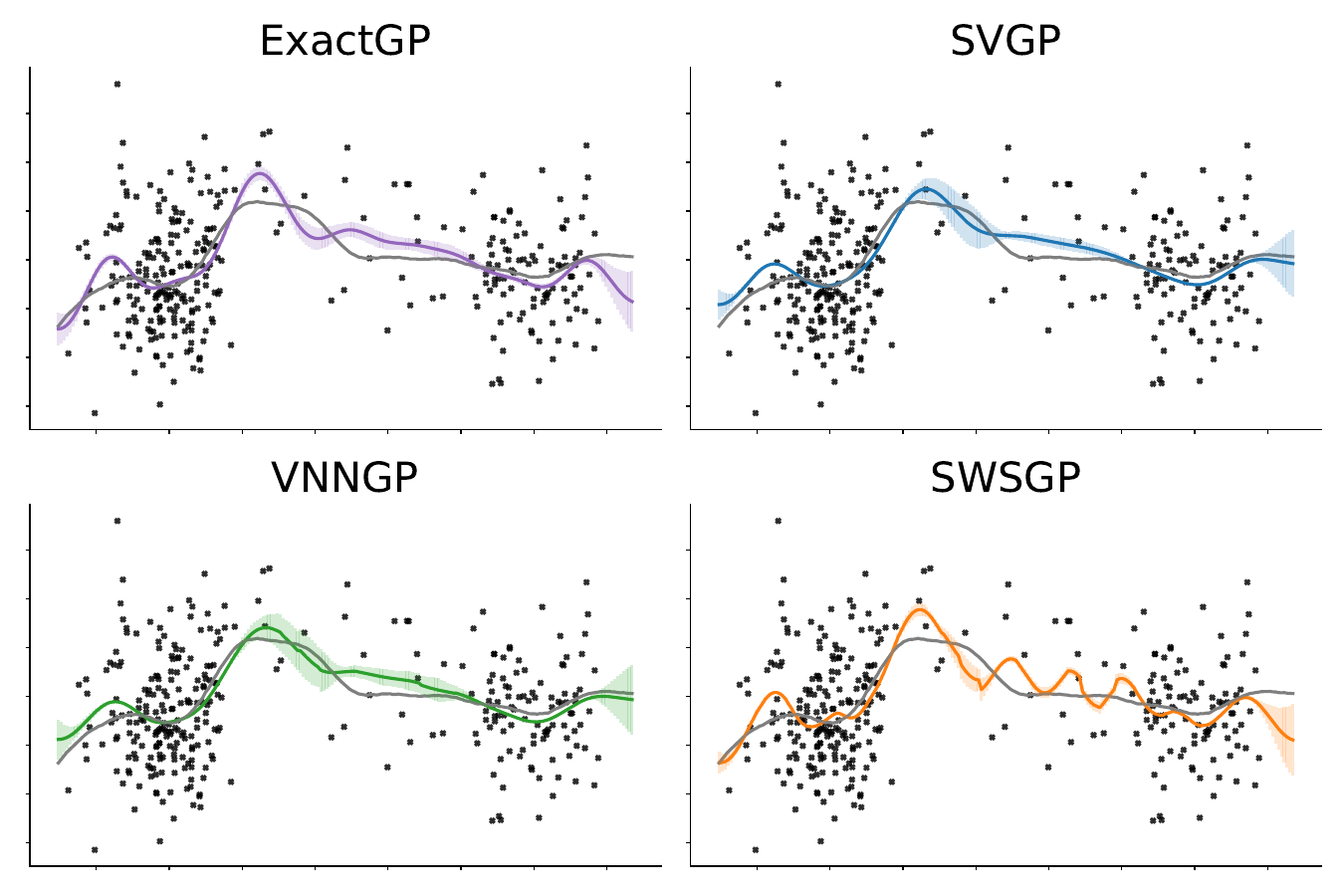}
\vspace{-1em}
 \caption{Posterior by Exact GP (purple), SVGP (blue), VNNGP (green) and SWSGP (orange), with 95\% uncertainty interval. The grey line is to the true data generating process, and scattering black dots are highly noisy observations.}
\label{fig:overfitting}
\end{center}
\vskip -0.2in
\end{figure}

In Fig~\ref{fig:overfitting}, we display the true data generating process, noisy observations and posteriors of all four models.
The SVGP and VNNGP posteriors are qualitatively close to that of exact GP, whereas SWSGP's posterior deviates from the underlying GP function in the data-sparse region. We suspect SWSGP's behavior to be due to its underestimating local KL divergence.
As we discuss in \cref{sec:compare-to-related-methods}, the ``local'' KL divergence of SWSGP will not regularize the variational distribution towards a globally smooth distribution. In particular, the SWSGP KL divergence concentrates on the data-dense regions, leading to a lack of regularization in the data-sparse region. By contrast, the VNNGP model is effectively regularized towards the GP prior throughout the input space.

\paragraph{Model selection.}
Finally, we turn to model selection by maximizing the ELBO.
Our goal is to see whether VNNGP or SWSGP is sensitive to  different settings of hyperparameters, in particular, the likelihood noise.  

\begin{figure}[!t]
\vskip 0.1in
\begin{center}
\includegraphics[scale=0.22]{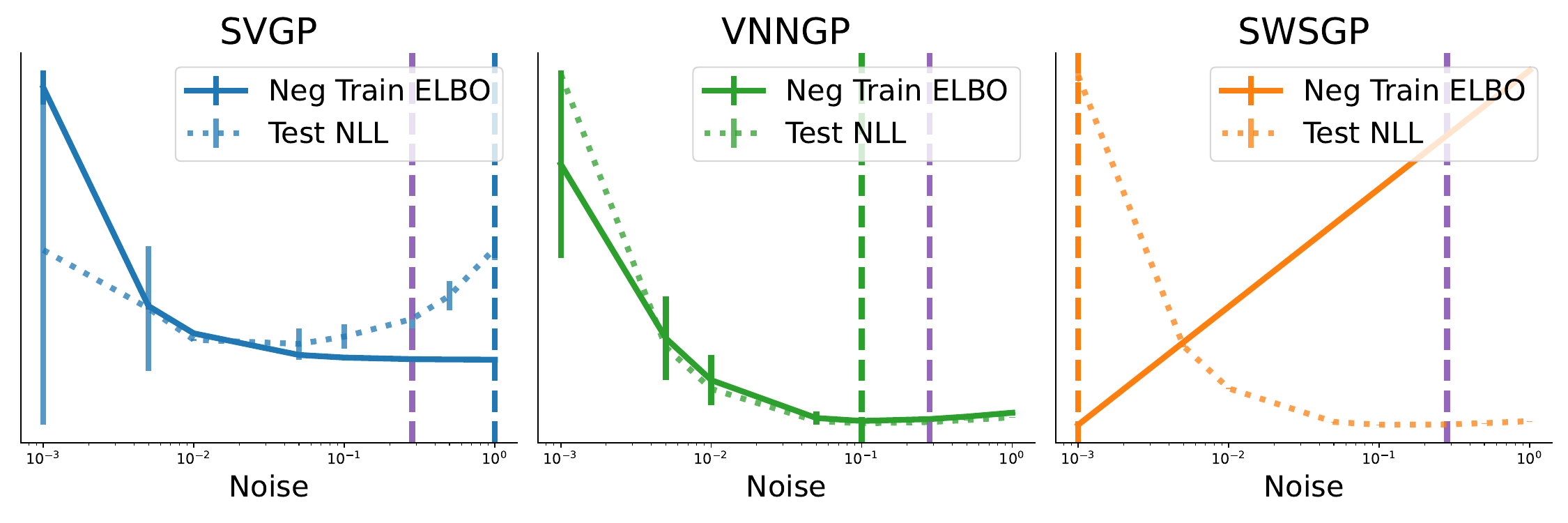}
\vspace{-2em}
 \caption{Negative training ELBO  (solid line) and test NLL (dotted line) as a function of the likelihood noise  by SVGP (blue), VNNGP (green) and SWSGP (orange) on UKHousing dataset. All ELBO and NLL values are scaled to $[0,1]$. The x-axis is in log-scale. The noise value picked by maximizing ELBO for each method is indicated by the vertical dashed line with corresponding color. The vertical purple line indicates the value ($0.28$) learned by exact GP.}
\label{fig:loss-vs-noise}
\end{center}
\vskip -0.25in
\end{figure}

We first fit exact GP to the UKHousing dataset under a Matern 5/2 kernel and a Gaussian likelihood.
We copy and fix the exact GP kernel hyperparameters to SVGP, VNNGP, and SWSGP models.
However, we vary the value of likelihood noise from 0.001 to 1.0.
We then optimize only the variational parameters of these three methods for 300 epochs. 
In Figure~\ref{fig:loss-vs-noise}, we plot the negative training ELBO and test NLL as a function of likelihood noise for each method.
We denote the noise parameters that maximize each method's training objective, as well as the noise parameter selected by the exact GP.
In agreement with prior work \citep{bauer2017understanding}, we find that SVGP tends to overestimate the noise parameter.
Both nearest neighbor methods tend to underestimate the likelihood noise.
However, the ELBO of SWSGP monotonically decreases as likelihood noise decreases, whereas the VNNGP ELBO achieves an optimum as a noise of $0.1$.
If we use the ELBO as a model selection criterion (as is often the case for variational methods), the resulting SWSGP model will have very low likelihood noise which results in poor test NLL.
Conversely, the VNNGP training ELBO is highly predictive of the test NLL, as both are roughly optimized as the same likelihood noise value.
We note that SWSGP's behavior in this setting is likely due to having $M = N$.
We repeat this experiment for SWSGP models that use $M \ll N$ inducing points in \cref{appendix:exp-model-selection},
and find that these models favor more reasonable likelihood noise values.
These observations suggest that the ELBO is a good model selection criteria for VNNGP;
however, ELBO optimization may result in extreme hyperparameters for SWSGP with large $M$.

\subsection{Predictive performance on real datasets}
\label{sec:exp-real-datasets}
\begin{table*}[!thbp]
\centering
\resizebox{0.9\linewidth}{!}{\begin{tabular}{ccccccccccc}
\toprule
        &        &  &  & \multicolumn{6}{c}{NLL} &  \\
        &        &  &  &           ExactGP &                          sgGP &                          SVGP &                       SWSGP-$M$ &                       SWSGP-$N$ &                         VNNGP &  \\
Dataset & $N$ & $D$ &          &                   &                               &                               &                               &                               &                               &          \\
\midrule
PoleTele & $9.6$K & $26$ &          &  $-.509 \pm .005$ &              $-.558 \pm .016$ &              $-.708 \pm .003$ &              $-.898 \pm .029$ &              $-.962 \pm .023$ &  $\mathbf{ -1.160 \pm .014 }$ &          \\
Elevators & $10.6$K & $16$ &          &   $.464 \pm .011$ &    $\mathbf{ .432 \pm .013 }$ &    $\mathbf{ .417 \pm .017 }$ &               $.691 \pm .049$ &               $.767 \pm .026$ &               $.463 \pm .020$ &          \\
Bike & $11.1$K & $17$ &          &  $-.744 \pm .007$ &           $53.364 \pm 27.690$ &  $\mathbf{ -1.849 \pm .016 }$ &             $-1.684 \pm .024$ &             $-1.405 \pm .096$ &             $-1.690 \pm .057$ &          \\
Kin40K & $25.6$K & $8$ &          &  $-.149 \pm .001$ &               $.701 \pm .139$ &              $-.399 \pm .003$ &              $-.747 \pm .016$ &  $\mathbf{ -1.010 \pm .005 }$ &  $\mathbf{ -1.016 \pm .004 }$ &          \\
Protein & $25.6$K & $9$ &          &  $1.044 \pm .003$ &               $.914 \pm .004$ &               $.967 \pm .005$ &               $.909 \pm .006$ &              $1.505 \pm .080$ &    $\mathbf{ .671 \pm .009 }$ &          \\
KEGG & $31.2$K & $20$ &          &  $-.713 \pm .004$ &  $\mathbf{ -1.013 \pm .035 }$ &             $-1.000 \pm .011$ &  $\mathbf{ -1.039 \pm .010 }$ &             $14.925 \pm .838$ &  $\mathbf{ -1.039 \pm .018 }$ &          \\
KEGGU & $40.7$K & $26$ &          &  $-.471 \pm .003$ &              $-.673 \pm .005$ &              $-.680 \pm .002$ &   $\mathbf{ -.719 \pm .004 }$ &             $41.675 \pm .427$ &   $\mathbf{ -.715 \pm .004 }$ &          \\
\hline
Precipitation & $64.8$K & $3$ &          &               --- &                           --- &               $.818 \pm .004$ &               $.396 \pm .001$ &              $1.809 \pm .224$ &    $\mathbf{ .145 \pm .002 }$ &          \\
UKHousing & $116$K & $2$ &          &   $.703 \pm .003$ &               $.598 \pm .001$ &               $.879 \pm .004$ &               $.455 \pm .003$ &              $1.050 \pm .013$ &    $\mathbf{ .367 \pm .003 }$ &          \\
3DRoad & $278$K & $3$ &          &   $.993 \pm .000$ &               $.712 \pm .001$ &               $.325 \pm .005$ &               $.584 \pm .003$ &  $\mathbf{ -1.476 \pm .079 }$ &              $-.048 \pm .406$ &          \\
Covtype & $372$K & $54$ &          &               --- &                           --- &               $.234 \pm .001$ &               $.172 \pm .001$ &    $\mathbf{ .069 \pm .000 }$ &               $.132 \pm .000$ &          \\
\bottomrule
\end{tabular}
}
\caption{Test set NLL (mean $\pm$ 1 standard error over 3 random seeds) on high-dimensional datasets (top 7 ones) and spatial datasets (bottom 4 ones). VNNGP achieves lowest NLL for most datasets. The result for RMSE is included in Table~\ref{tab:uci-prediction-rmse} in appendix.}
\label{tab:uci-prediction-nll}
\end{table*}
We conduct an extensive evaluation of our method on real datasets. 
We consider a wide range of high dimensional and spatiotemporal datasets from the UCI repository \citep{asuncion2007uci}. In addition we include three spatial datasdets, UKHousing as mentioned in Section~\ref{sec:exp-hyperparameter-sensitivity}, Precipitation (a monthly precipitation dataset with $D=3$) \citep{lyon2004strength, lyon2005enso} and Covtype (a tree cover dataset with $D=54$) \footnote{The task for Covtype is predicting whether the primary tree cover at a given location is pine trees or other types of trees. Despite this dataset is not 2 or 3-dimensional, its features essentially codify location information. Therefore we categorize Covtype as a spatial dataset.}  \citep{blackard1999comparative}.    
Our goal is primarily to compare VNNGP to other variational methods;
however, we include the additional baselines of {\bf Exact GP}---utilizing the methodology of \citet{wang2019exact})---%
and {\bf sgGP} \citep{chen2020stochastic}---which is a non-variational nearest neighbor benchmark.
We do not compare to the Amortized Inference GP since it is not amenable to kernel learning. For VNNGP and SWSGP, we tune the number of nearest neighbors $K$ in $\{32, 256\}$. Additionally, we include two variants of the SWSGP method: with SWSGP-$N$ we set all inducing points at observed locations, and with SWSGP-$M$ we allow using  $M<N$ inducing points -- the latter is what \citet{tran2021sparse} originally experimented with. In particular for SWSGP-$M$, we recompute nearest neighbors every epoch and we tune $M$ in $\{1024, 2048, 4096, 8192\}$. For sgGP, the subsampled datasets are constructed by selecting a random point $\bx, y$ and its 15 nearest neighbors as in \citet{chen2020stochastic}.  Each dataset is randomly split to 64\% training, 16\% validation and 20\% testing sets. Precipitation dataset uses a student-t likelihood and Covtype dataset uses a bernoulli likelihood, otherwise a Gaussian likelihood is used. 
All kernels are Matern 5/2 with a
separate lengthscale per dimension. We run each model with three random seeds. For each random seed, VNNGP applies different random ordering of inducing points. 
See appendix \ref{appendix:training-details} for more details. 

We report test NLL in Table~\ref{tab:uci-prediction-nll}, and the result for test RMSE is included in Table~\ref{tab:uci-prediction-rmse} of appendix. From the tables, we make three key comparisons. (i) Exact GP and sgGP cannot run on problems with non-Gaussian likelihoods (Precipitation and Covtype) while variational methods are still viable. (ii) VNNGP substaintially outperforms other nearest neighbor methods. Specifically, we note that sgGP and SWSGP-$N$ somtimes obtain a very high NLL, while  SWSGP-$M$ does not suffer from this issue. We suspect this result is attributed to that sgGP and SWSGP-$N$ tend to learn a very small likelihood noise on certain datasets (we report the learned noise in Table~\ref{tab:uci-prediction-noise} of appendix); Also recall from Section~\ref{sec:exp-comparison-to-swsgp} that  maximizing the ELBO of SWSGP-$N$ leads to extremely small noise values. (iii) While there are a few datasets (Elevators and Bike) where SVGP excels, VNNGP outperforms SVGP especially on large-scale,  spatial datasets. SVGP's performance is limited by using $M=1024 \ll N$ inducing points, which is detrimental on these problems, while VNNGP has the ability to scale $M$ to $N$ with a sparse approximation. 

\begin{figure}[!t]
    \centering
    \includegraphics[scale=0.25]{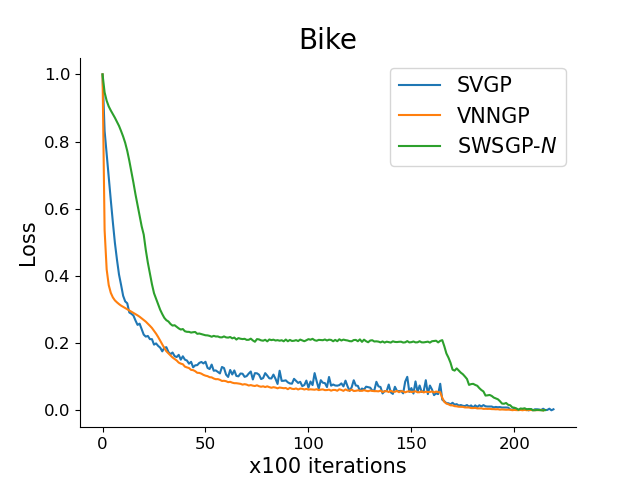}
    \includegraphics[scale=0.25]{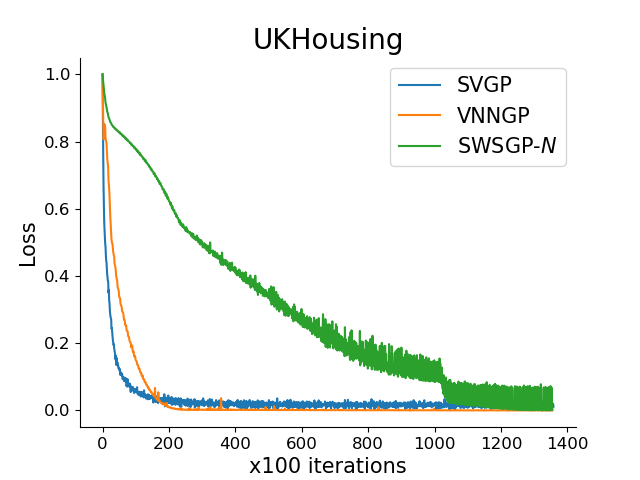}
    \vspace{-2em}
    \caption{Training loss versus iterations on Bike and UKhousing. The training loss is averaged over every 100 iterations and all loss values are standardized. Despite VNNGP has more parameters, its convergence speed is comparable to SVGP.}
    \vskip -0.2in
    \label{fig:exp-convergence-speed}
\end{figure}
Finally, we note that both VNNGP and SWSGP-$N$ use much more parameters than SVGP. It is natural to ask whether this fact will impede their convergence speeds. In Figure~\ref{fig:exp-convergence-speed}, we plot the training loss versus number of training iterations for SVGP, VNNGP and SWSGP-$N$ on Bike and UKHousing.  We observe that all three methods reach convergence by the end of training. Moreover, VNNGP's convergence speed is comparable to SVGP. This result demonstrates that the larger number of model parameters does not hurt VNNGP's optimization process, making it a practical model to use.  

\section{Discussion}
In this work, we propose VNNGP, a scalable GP method that combines a sparse precision approximation with variational inference. 
This sparse approximation allows VNNGP to use orders of magnitude more inducing points than other variational methods. We perform an extensive empirical evaluation and show that VNNGP obtains strong performance compared to other baselines. 

We consider two extensions for future works. Similar to \citet{liu2019amortized} and \citet{jafrasteh2021input}, we can apply amortized learning to variational parameters.
This could greatly reduce the number of parameters and potentially accelerate optimization. Another extension is to select nearest neighbors using metrics other than Euclidean distance.
For example, \citet{kang2021correlation} suggest that the prior covariance function can be a metric to select the nearest neighbor set. This could enhance VNNGP's accuracy on problems with anisotropic or periodic covariance structure.

\bibliography{citation}
\bibliographystyle{icml2022}

\newpage
\appendix
\onecolumn

\section{Computing the ELBO for VNNGP}
\label{appendix:vnngp-elbo-computation}
In this section, we derive the exact computation of the ELBO for VNNGP and analyze the computational complexity. 

Following Eq~\ref{eq:vnngp-elbo}, and using the derivations from Eq~\ref{eq:data-likelihood-term} and Eq~\ref{eq:vnngp-kl-factorization}, we have 
 \begin{align}
\mathcal{L}_{\textrm{VNNGP}} &= \underbrace{  \sum_{i=1}^N \mathbb{E}_{q(\bfn, \bu)} [\log p(y_i | f_i)]}_{\textrm{data likehood}}  - \underbrace{\kl{q(\bu)}{p(\bu)}}_{\textrm{KL divergence}},  \\
&= \underbrace{\sum_{i=1}^N \int q(f_i) \log p(y_i | f_i) d f_i}_{\textrm{data likelihood}} -  \underbrace{\sum_{j=1}^M \mathbb{E}_{q(\bu_{n(j)})} \left[ \kl{q(u_j)}{p(u_j | \bu_{n(j)})} \right]}_{\textrm{kl divergence}}. 
\label{eq:appendix-vnngp-elbo}
\end{align}
where
\begin{align}
    q(f_i) &= \mathbb{E}_{q(\bu)} \left[ p(f_i | \bu_{n(i)}) \right] \\
    &= \mathcal{N} (f_i | \bk_{n(i), i}^\top \bK_{n(i), n(i)}^{-1} \bmm_{n(i)}, k_{i,i} - \bk_{n(i), i}^\top \bK_{n(i), n(i)}^{-1} \bk_{n(i), i}  +  \bk_{n(i), i}^\top \bK_{n(i), n(i)}^{-1}\bS_{n(i)}  \bK_{n(i), n(i)}^{-1} \bk_{n(i), i}),
    \label{eq:appendix-qf}
\end{align}
and  $\bS_{n(i)}$ is a diagonal matrix with diagonal elements being $\bs_{n(i)}$.

\begin{itemize}
    \item The data likelihood term is tractable for Gaussian likelihood; otherwise, techniques such as MCMC sampling are required. The major complexity in evaluating each expected likelihood term comes from computing $q(f_i)$, which essentially resides at inverting the $K\times K$ matrix $\bK_{n(i), n(i)}$ which takes $O(K^3)$ time. 

\item To compute the KL divergence term,  we first compute each conditional KL term. Given fixed $\bu_{n(j)}$, 
\begin{align}
    \kl{q(u_j)}{p(u_j | \bu_j)} &= \kl{\mathcal{N} (u_j | m_j, s_j)}{\mathcal{N} (u_j | \bb_j^\top \bu_{n(j)}, f_j)} \\
    &=\frac{1}{2} \left[ \log f_j - \log s_j - 1 + f_j^{-1} s_j +f_j^{-1}  (m_j - \bb_j^\top \bu_{n(j)})^2  \right],
\end{align}
where $ f_j = k_{j,j}-\bk^\top_{n(j),j} \bK_{n(j), n(j)}^{-1} \bk_{n(j),j}$ and 
 $\bb_j =  \bK_{n(j), n(j)}^{-1}\bk_{n(j),j}$. Integrating over $q(\bu_{n(j)})$ and summing over inducing points we obtain, 
\begin{align}
    \textrm{kl divergence} &=  \frac{1}{2} \sum_{j=1}^M  \log f_j - \log s_j - 1 + f_j^{-1} \left[ s_j + (\bb_j^2)^\top \bs_{n(j)} + (m_j - \bb_j^\top \bmm_{n(j)})^2 \right], 
\end{align}
where $\bb_j^2$ denotes the element-wise squared of vector $\bb_j$. Computing each summation term in the KL divergence requires inverting an $K \times K$ matrix, i.e. $\bK_{n(j), n(j)}$, which is $O(K^3)$ complexity.  
\end{itemize}

Therefore, the overall complexity of estimating the ELBO using a mini-batch of $N_b$ data points and a mini-batch of $M_b$ inducing points is $O(N_b K^3 + M_b K^3)$. 

\section{Extension to more expressive variational family}
\label{appendix:sec-variational-family-extension}
We now consider more expressive variational approximations  and discuss how to mini-batch the VNNGP's ELBO computation. We assume the following variational family: 
\begin{align}
        q(\bu)  &= \mathcal{N} (\bu | \bmm, \tilde \bL \tilde \bL^\top)  \\ 
        q(\bfn | \bu) &= \prod_{i=1}^N p(f_i | \bu_{n(i)})
\end{align}
where $\tilde \bL$ is the lower-triangular Cholesky factor for variational posterior covariance (potentially full-rank). The mean-field case discussed in appendix  \ref{appendix:vnngp-elbo-computation} is a special case where we make $\tilde L$ diagonal. 

Similarly, the ELBO under this variational family decomposes into the data likelihood term and the KL divergence term as in Eq~\ref{eq:appendix-vnngp-elbo} with the modification that  
\begin{align}
    q(f_i) &= \mathbb{E}_{q(\bu)} \left[ p(f_i | \bu_{n(i)}) \right] \\
    &= \mathcal{N} (f_i | \bk_{n(i), i}^\top \bK_{n(i), n(i)}^{-1} \bmm_{n(i)}, k_{i,i} - \bk_{n(i), i}^\top \bK_{n(i), n(i)}^{-1} \bk_{n(i), i}  +  \bk_{n(i), i}^\top \tilde{\bL}_{n(i)} \tilde{\bL}_{n(i)}^\top  \bk_{n(i), i}). 
    \label{eq:appendix-qf-2}
\end{align}

\begin{itemize}
    \item The likelihood computation is similar to the mean-field case in appendix  \ref{appendix:vnngp-elbo-computation}, except that we are now integrating over $q(f_i)$ in Eq~\ref{eq:appendix-qf-2} instead of $q(f_i)$ in Eq~\ref{eq:appendix-qf}. The former involves a potentially full-rank $K \times K$ variational covariance $\tilde{\bL}_{n(i)} \tilde{\bL}_{n(i)}^\top $ while the latter involves a diagonal one.  

\item KL divergence. 

For mathematical convenicence, we consider the Cholesky composition of prior precision for inducing points $\bK_{\bz, \bz}^{-1} = \bL^\top \bL$ where $\bL$ is a lower triangular matrix. Each row of $\bL$ could be separately computed at $O(K^3)$ cost (see \citet{datta2021nearest} for details). The KL divergence can be computed as follows: 
\begin{align}
        \kl{q(\bu)}{p(\bu)} &= \kl{\mathcal{N}(\bmm, \tilde \bL \tilde \bL^\top)}{\mathcal{N} (0, (\bL^\top \bL)^{-1} )} \\
        &= \frac{1}{2} [-2 \log | \bL| - 2 \log | \tilde \bL |  - M + \textrm{tr} (\bL^\top \bL \tilde \bL \tilde \bL^\top) + \bmm^\top \bL^\top \bL \bmm ]
    \end{align}

 Here we take a special treatment of the trace term:  
    \begin{align}
        \textrm{tr} (\bL^\top \bL \tilde \bL \tilde \bL^\top) &= \textrm{tr} (\bL \tilde \bL (\bL \tilde \bL)^\top  ) \textrm{ (by cyclic property of trace)} \\
        &= \sum_{i=1}^M \sum_{j=1}^M (\bL \tilde \bL)^2_{ij} \textrm{  (by definition of trace)}\\ 
        &= \sum_{i=1}^M \sum_{j=1}^M \sum_{k=1}^M L^2_{ij} \tilde L^2_{jk}  \textrm{  (expanding the $(\bL \tilde \bL_{ij})$ terms)}\\ 
        &= \sum_{i=1}^M \sum_{j=1}^M \sum_{k \in n(i) \cup \{i\}} \bL_{ik}^2 \tilde \bL_{kj}^2 \textrm{ (using row-sparsity of $\bL$)} 
        \end{align} 
Now we re-write the KL by decomposing each term: 
\begin{align}
        \kl{q(\bu)}{p(\bu)} &= \frac{1}{2} [-2 \sum_{i=1}^M \log L_{ii} - 2 \sum_{i=1}^M \tilde L_{ii}  - M + \sum_{i=1}^M \sum_{j=1}^M \sum_{k \in n(i) \cup \{i\}} \bL_{ik}^2 \tilde \bL_{kj}^2  + \sum_{i=1}^M (\sum_{k \in n(i) \cup \{i\}} L_{ik} m_k)^2 ] \\
        &= \frac{1}{2} \sum_{i=1}^M [-2 \log L_{ii} - 2 \tilde L_{ii}  - 1 + \sum_{k \in n(i) \cup \{i\}} \bL_{ik}^2 \sum_{j=1}^M \tilde \bL_{kj}^2  + (\sum_{k \in n(i) \cup \{i\}} L_{ik} m_k)^2 ], 
    \end{align}
    which again factorizes over inducing points $i=1:M$. However, one should note that computing each KL term now scales $O(MK^3)$ due to the complexity in trace term $\sum_{k \in n(i) \cup \{i\}} \sum_{j=1}^M  \bL_{ik}^2 \tilde \bL_{kj}^2$. One can alleviate this issue by (i) obtaining a stochastic estimate for the trace term (i.e. sub-sample $j$) (ii) introduce sparsity structure in $\tilde \bL$ / variational posterior. 
\end{itemize}
In summary, we explore the possibility of extending VNNGP to more expressive variational family. We show that under the full-rank variational family, the VNNGP's ELBO still factorizes over data points and inducing points. Computing each term now takes more computations and but techniques can be applied to improve the scalability.

\section{KL divergence for SWSGP}
\label{appendix:swsgp-kl-divergence}
Here we re-state and prove the two claims about KL divergence for SWSGP in Section~\ref{sec:compare-to-related-methods}. 

Consider a fixed prior $p(\bu)$ and a fixed variational posterior $q(\bu)$ over $M$ inducing points $\bu$. Then we claim that 
\begin{enumerate}
    \item  SWSGP's KL divergence is always underestimating, if not equal to, SVGP's KL divergence. That is, for any subset $\mathcal{S} \in \{1, 2, \cdots, M\}$.  
    \begin{align}
        \kl{q(\bu_{\mathcal{S}})}{p(\bu_{\mathcal{S}})} & \leq \kl{q(\bu)}{p(\bu)}, 
    \end{align}
    \item The amount of underestimation depends on the size of the nearest neighbor set. More precisely, for any two  subsets $\mathcal{S}_1$ and  $\mathcal{S}_2$ such that $\mathcal{S}_1 \subset \mathcal{S}_2 \subset \{ 1, \cdots, M\}$,  
    \begin{align}
         \kl{q(\bu_{\mathcal{S}_1})}{p(\bu_{\mathcal{S}_1})} & \leq\kl{q(\bu_{\mathcal{S}_1})}{p(\bu_{\mathcal{S}_2})}. 
    \end{align}
\end{enumerate}
The proof for the above claims will immediately follow once we prove the following proposition. 

\begin{proposition}\label{prop:swsgp-kl-bound}
Let $p(\bu)$ and $q(\bu)$ be two distributions for an $M$-dimensional random variable $\bu$, and let $\bu_1$ be any sub-vector of $\bu$, then 
\begin{align}
    \kl{q(\bu_1)}{p(\bu_1)} & \leq
    \kl{q(\bu)}{p(\bu)}. 
     \label{eq:swsgp-kl-inequality}
\end{align}

\end{proposition}

\begin{proof} The proof relies on the chain rule of probability distribution, and the fact that the KL divergence is non-negative. 

Denote $\bu_2$ as the remaining sub-vector by taking $\bu_1$ out of $\bu$. It follows that 
\begin{align}
    & \kl{q(\bu )}{ p(\bu)} \\
    &= \int q(\bu) \log \frac{q(\bu )}{p(\bu ) } d\bu \\
    &= \int \int q(\bu_1) q(\bu_2 | \bu_1) \log \frac{q(\bu_1) q(\bu_2 | \bu_1)}{p(\bu_1) p(\bu_2 | \bu_1)} d\bu_1 d\bu_2 \\
    &= I + II, 
\end{align}
where 
\begin{align}
    I & \equiv \int q(\bu_1) \log \frac{q(\bu_1)}{p(\bu_1)} \left( \int  q(\bu_2 | \bu_1 )  d \bu_2 \right) d \bu_1 = \kl{ q(\bu_1) }{ p(\bu_1) }, \\
    II & \equiv \int q(\bu_1) \left( \int q(\bu_2 | \bu_1 ) \log \frac{q(\bu_2 | \bu_1)}{p(\bu_2 | \bu_1 )} \right) d \bu_1 = \int q(\bu_1) \kl{q(\bu_2 | \bu_1)}{ p(\bu_2 | \bu_1)} d\bu_1. 
\end{align}
Note that for each $\bu_1$ fixed, $ \kl{q(\bu_2 | \bu_1)}{ p(\bu_2 | \bu_1)}$ defines a (conditional) KL divergence and is non-negative. Therefore, $\kl{q(\bu)}{p(\bu )} = I + II \geq I$, i.e.   Eq~\ref{eq:swsgp-kl-inequality} holds. 
\end{proof}

\section{Experiment details and additional results}

\begin{figure*}[!thp]
\vskip 0.1in
\begin{center}
\includegraphics[scale=0.4]{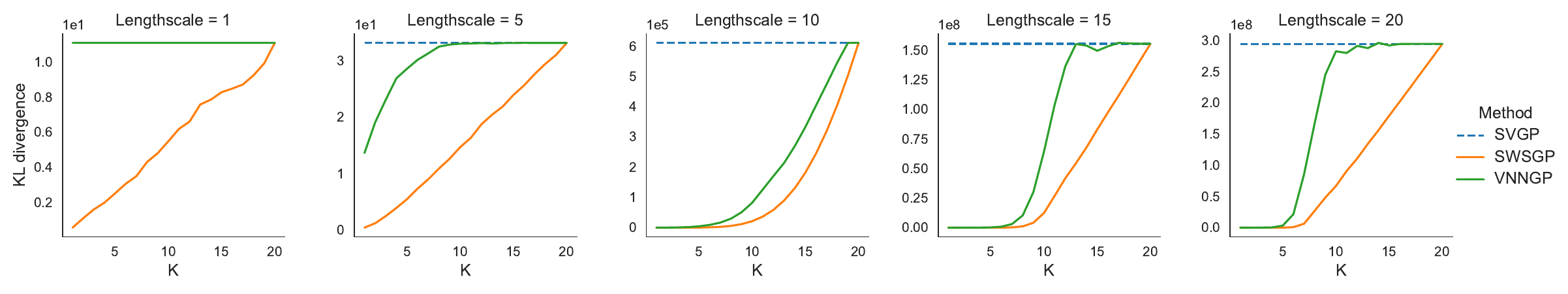}
\includegraphics[scale=0.4]{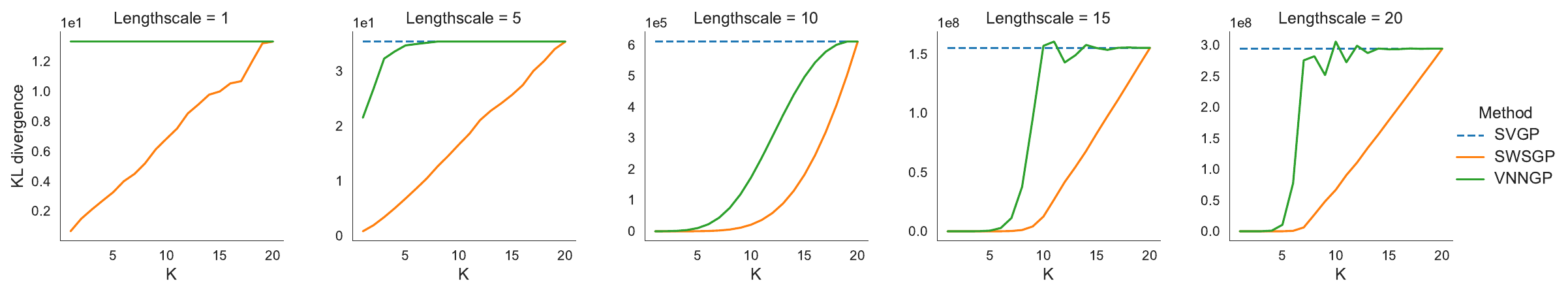}
\includegraphics[scale=0.4]{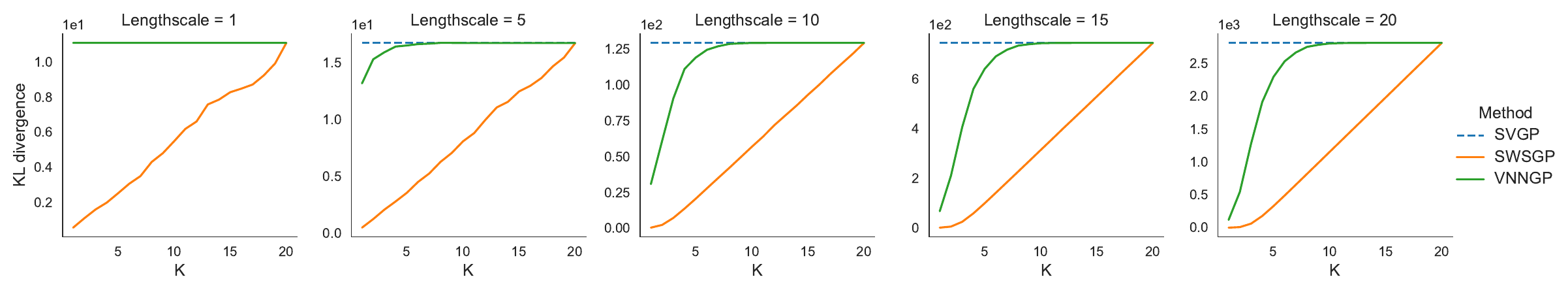}
\includegraphics[scale=0.3]{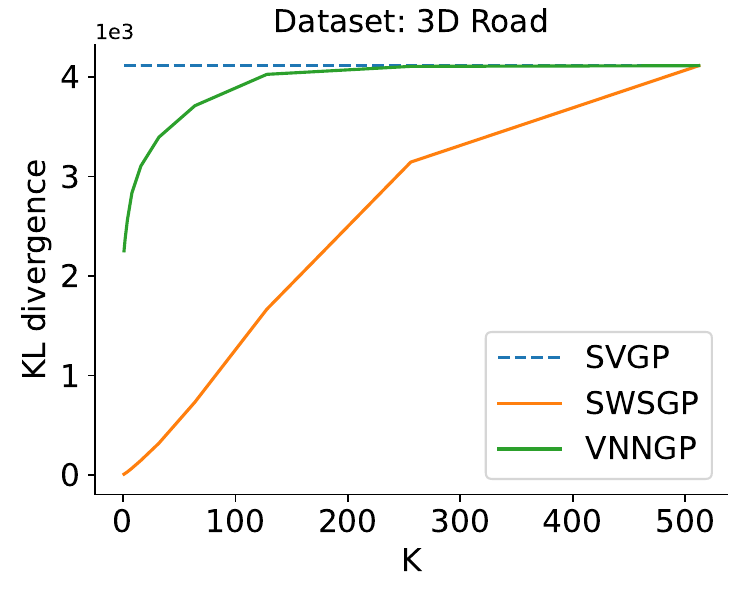}
\caption{KL comparison under different settings than the setting of  Figure~\ref{fig:kl}. Row 1: A different random ordering of induing points. 
Row 2: Coordinate ordering of inducing points. Row 3: Matern (2.5) kernel. Row 4: Real dataset (3DRoad). } 
\label{fig:kl-additional}
\end{center}
\vskip -0.2in
\end{figure*}

\subsection{KL divergence comparison}
\label{appendix:exp-kl-divergence}
In Figure~\ref{fig:kl-additional}, we vary the setting in Section~\ref{sec:exp-comparison-to-swsgp}: different orderings of inducing points, different kernel functions, and real data v.s. synthetic data. From this figure, we observe similar patterns as from Figure~\ref{fig:kl}.

\subsection{Model selection} 
\label{appendix:exp-model-selection}
We include additional results for the model selection experiment in Section~\ref{sec:exp-comparison-to-swsgp}. We refer the models in Section~\ref{sec:exp-comparison-to-swsgp} by VNNGP-$N$ and SWSGP-$N$ that set inducing points on observed locations. Here, we consider the alternative versions, VNNGP-$M$ and SWSGP-$M$, which use $M \ll N$ inducing points for $M \in \{ 1024, 8192\}$, and the inducing point locations are learned during optimization. Correspondingly, the nearest neighbors are updated every training epoch for VNNGP and SWSGP.  The other configurations are the same as Section~\ref{sec:exp-comparison-to-swsgp} (especially,  SVGP uses $1024$ inducing points for all times). All inducing point locations are initialized by k-means clustering of observed locations. 

\begin{figure}[!htp]
     \centering
         \includegraphics[scale=0.26]{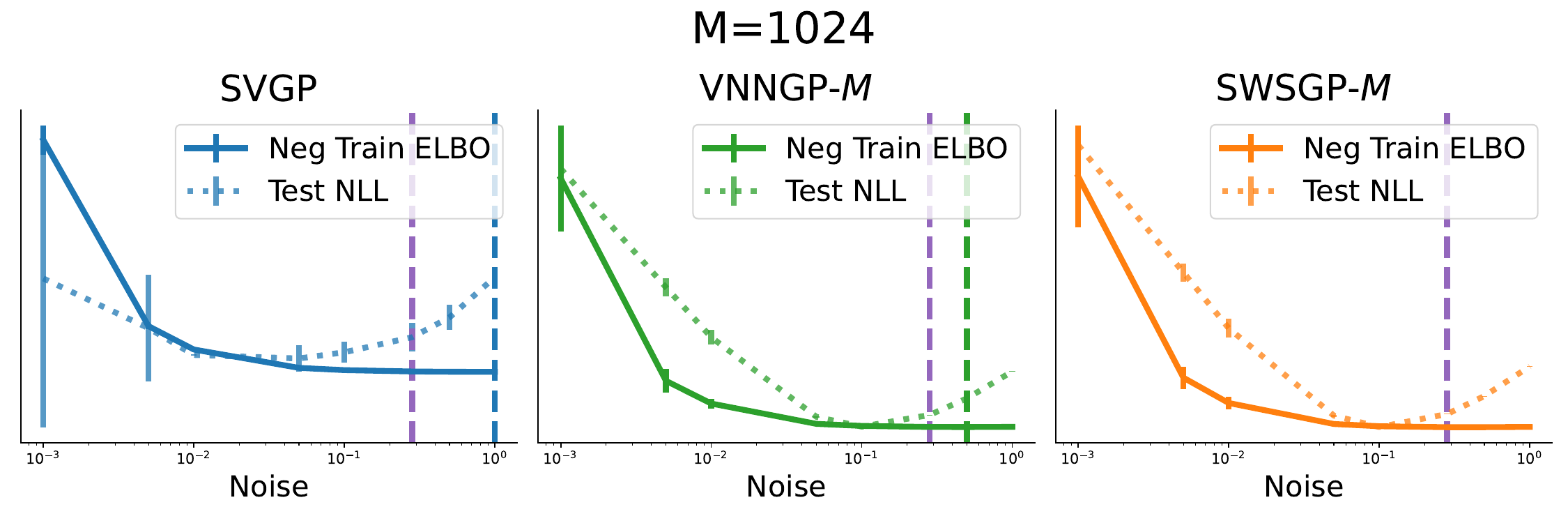}
       \includegraphics[scale=0.26]{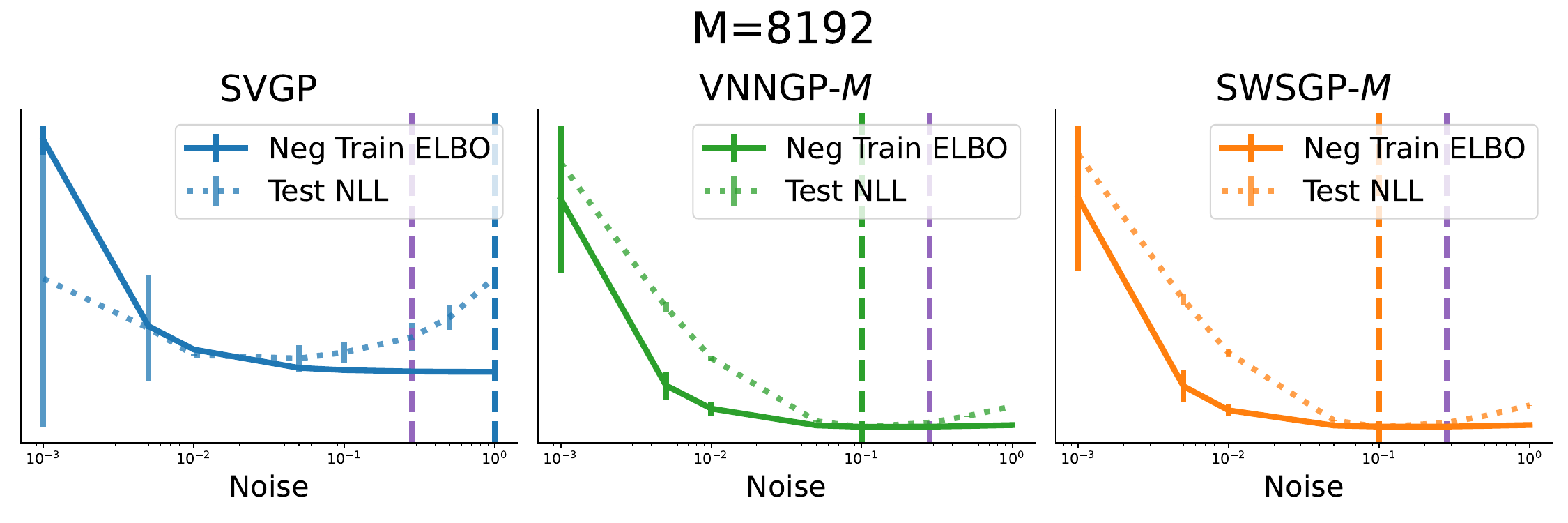}
        \caption{Negative ELBO in the last training iteration (solid line) and test set NLL (dotted line) as a function of the likelihood noise parameter by SVGP (blue), VNNGP (green) and SWSGP (orange) on UKHousing dataset. All ELBO and NLL values are scaled to $[0,1]$.  
        The x-axis (Noise) is in log-scale. The best noise parameter picked by maximizing ELBO for each method is indicated by the vertical dashed line with  corresponding color. The vertical purple line indicates the noise parameter ($=0.28$) learned by exact GP. 
        \textbf{Left}: both VNNGP and SWSGP use $M=1024$ inducing points. \textbf{Right}: both VNNGP and SWSGP use $M=8192$ inducing point. For all cases, SVGP use $1024$ inducing points. }
        \label{fig:loss-vs-noise-additional}
\end{figure}

Comparing Figure~\ref{fig:loss-vs-noise-additional} to Figure~\ref{fig:loss-vs-noise}, we can see that (1) VNNGP-$M$ and VNNGP-$N$ have consistent behaviors in model selection. (2) While maximizing the ELBO of SWSGP-$N$ tends to favor a model with small likelihood noise, this is not the case for SWSGP-$M$. As the noise parameter becomes extremely small, the training ELBO of SWSGP-$M$ decreases. As a result, optimizing SWSGP-$M$ leads to a model that has a ``medium'' likelihood noise and a low test NLL. These observations suggest that maximizing the ELBO for SWSGP-$N$ (and potentially for SWSGP-$M$ with $M$ close to $N$) can lead to undesired model selection outcome, while this does not hold for SWSGP-$M$ with a medium size of inducing points.

\subsection{Predictive performance on real dataset}
Here we provide training details and additional results for Section~\ref{sec:exp-real-datasets}.

\subsubsection{Training details} 
\label{appendix:training-details}
For all methods, we use an Adam optimizer and a MultiStepLR scheduler
dropping the learning rate by a factor of 10 at the 75\% and 90\% of the optimization iterations; all kernels are Matern 5/2 kernel with separate lengthscale per dimension;
the kernel lengthscales, outputscale and likelihood noise parameter (if any) are all initialized as $0.6931$, except that exact GP is initiliazed with lengthscale $=0.01$ on 3DRoad dataset. All hyperparameters are picked by the best validation NLL. 

All methods are trained with $\{300, 500\}$ iterations and learning rate of $\{0.005, 0.01\}$ for datasets of size below 50$K$, and with $\{100, 300\}$ iterations and learning rate of $\{0.005, 0.001\}$ for above 50$K$.

For sgGP, we train using minibatches of 16 data points. As suggested by \citet{chen2020stochastic}, the minibatches are constructed by sampling one training data point and selecting
its 15 nearest neighbors. To accelerate optimization, we accumulate the gradients of 1,024 minibatches before per-
forming an optimization step (these 1,024 minibatch updates can be performed in parallel, enabling GPU acceleration). 

For SVGP, we use $M=1024$ inducing points and a full-rank variational approximation. The inducing point locations are initilaized by k-means clustering of observed locations

VNNGP, SWSGP-$M$ and SWSGP-$N$ use a mean-field variational approximation. SWSGP-$M$ chooses $M\in \{1024, 2048, 4096, 8192\}$, and the inducing point locations are initilaized by k-means clustering of observed locations. All three methods tune $K$ in $\{ 32, 256\}$. 

\subsubsection{Predictive RMSE table} 
The test set RMSE results for reported in Table~\ref{tab:uci-prediction-rmse}. From the table, we observe that VNNGP obtains comparable RMSE values to other state-of-the-arts and excels at some datasets. 

\begin{table*}[!thbp]
\centering
\resizebox{0.9\linewidth}{!}{\begin{tabular}{ccccccccccc}
\toprule
        &        &  &  & \multicolumn{6}{c}{RMSE} &  \\
        &        &  &  &                     ExactGP &                        sgGP &                        SVGP &                     SWSGP-$M$ &                     SWSGP-$N$ &                       VNNGP &  \\
Dataset & $N$ & $D$ &          &                             &                             &                             &                             &                             &                             &          \\
\midrule
PoleTele & $9.6$K & $26$ &          &             $.095 \pm .000$ &  $\mathbf{ .092 \pm .002 }$ &             $.113 \pm .000$ &             $.101 \pm .002$ &             $.098 \pm .003$ &  $\mathbf{ .091 \pm .002 }$ &          \\
Elevators & $10.6$K & $16$ &          &  $\mathbf{ .360 \pm .006 }$ &  $\mathbf{ .358 \pm .006 }$ &  $\mathbf{ .366 \pm .006 }$ &             $.399 \pm .008$ &             $.456 \pm .001$ &             $.373 \pm .004$ &          \\
Bike & $11.1$K & $17$ &          &             $.054 \pm .003$ &             $.400 \pm .153$ &  $\mathbf{ .028 \pm .001 }$ &             $.037 \pm .002$ &             $.042 \pm .006$ &             $.043 \pm .003$ &          \\
Kin40K & $25.6$K & $8$ &          &             $.099 \pm .001$ &             $.458 \pm .076$ &             $.145 \pm .001$ &             $.118 \pm .002$ &  $\mathbf{ .097 \pm .001 }$ &  $\mathbf{ .096 \pm .001 }$ &          \\
Protein & $25.6$K & $9$ &          &  $\mathbf{ .533 \pm .002 }$ &  $\mathbf{ .534 \pm .006 }$ &             $.633 \pm .003$ &             $.600 \pm .004$ &             $.592 \pm .005$ &             $.565 \pm .003$ &          \\
KEGG & $31.2$K & $20$ &          &  $\mathbf{ .085 \pm .002 }$ &  $\mathbf{ .087 \pm .003 }$ &             $.089 \pm .001$ &  $\mathbf{ .086 \pm .001 }$ &  $\mathbf{ .085 \pm .002 }$ &  $\mathbf{ .085 \pm .001 }$ &          \\
KEGGU & $40.7$K & $26$ &          &  $\mathbf{ .117 \pm .000 }$ &             $.122 \pm .001$ &             $.122 \pm .000$ &  $\mathbf{ .118 \pm .001 }$ &  $\mathbf{ .116 \pm .001 }$ &             $.118 \pm .000$ &          \\
\hline
Precipitation & $64.8$K & $3$ &          &                         --- &                         --- &             $.635 \pm .004$ &             $.485 \pm .001$ &             $.506 \pm .001$ &  $\mathbf{ .432 \pm .002 }$ &          \\
UKHousing & $116$K & $2$ &          &             $.402 \pm .001$ &             $.359 \pm .001$ &             $.582 \pm .003$ &             $.382 \pm .001$ &             $.498 \pm .000$ &  $\mathbf{ .346 \pm .002 }$ &          \\
3DRoad & $278$K & $3$ &          &             $.165 \pm .000$ &             $.187 \pm .001$ &             $.329 \pm .001$ &             $.432 \pm .001$ &  $\mathbf{ .097 \pm .017 }$ &             $.298 \pm .064$ &          \\
Covtype & $372$K & $54$ &          &                         --- &                         --- &             $.671 \pm .001$ &  $\mathbf{ .657 \pm .001 }$ &             $.670 \pm .001$ &             $.671 \pm .001$ &          \\
\bottomrule
\end{tabular}
}
\caption{Test set RMSE (mean $\pm$ 1 standard error over 3 random seeds) on high-dimensional datasets (top 8 ones) and spatial datasets (bottom 4 ones). }
\label{tab:uci-prediction-rmse}
\end{table*}

\subsubsection{Learned likelihood noise table} 

We report the learned value of likelihood noise for all models in Table~\ref{tab:uci-prediction-noise}. Note that Covtype dataset uses Bernoulli likelihood which does not have noise parameter,  therefore it is not included. 

We observe in Table~\ref{tab:uci-prediction-nll} that sgGP and SWSGP-$N$ can sometimes obtain a very high NLL values. We suspect that is due to they learn a particularly small likelihood noise on corresponding datasets. For example, sgGP's NLL is approximately 53 on Bike, and its learned noise is around 0; SWSGP-$N$'s NLL is approximately 41 on KEGGU, and its learned noise is around 0. 

\begin{table*}[!thbp]
\centering
\resizebox{0.9\linewidth}{!}{\begin{tabular}{ccccccccccc}
\toprule
        &        &  &  & \multicolumn{6}{c}{noise} &  \\
        &        &  &  &          ExactGP &                        sgGP &               SVGP &            SWSGP-$M$ &                     SWSGP-$N$ &              VNNGP &  \\
Dataset & $N$ & $D$ &          &                  &                             &                    &                    &                             &                    &          \\
\midrule
PoleTele & $9.6$K & $26$ &          &  $.020 \pm .000$ &             $.002 \pm .000$ &    $.015 \pm .000$ &    $.005 \pm .000$ &  $ { .000 \pm .000 }$ &    $.002 \pm .000$ &          \\
Elevators & $10.6$K & $16$ &          &  $.166 \pm .001$ &             $.102 \pm .001$ &    $.132 \pm .001$ &    $.062 \pm .000$ &  $ { .000 \pm .000 }$ &    $.067 \pm .000$ &          \\
Bike & $11.1$K & $17$ &          &  $.016 \pm .000$ &  $ { .000 \pm .000 }$ &    $.002 \pm .000$ &    $.002 \pm .000$ &             $.000 \pm .000$ &    $.001 \pm .000$ &          \\
Kin40K & $25.6$K & $8$ &          &  $.020 \pm .000$ &  $ { .000 \pm .000 }$ &    $.031 \pm .000$ &    $.009 \pm .000$ &             $.000 \pm .000$ &    $.001 \pm .000$ &          \\
Protein & $25.6$K & $9$ &          &  $.182 \pm .003$ &             $.022 \pm .002$ &    $.417 \pm .001$ &    $.339 \pm .001$ &  $ { .000 \pm .000 }$ &    $.018 \pm .000$ &          \\
KEGG & $31.2$K & $20$ &          &  $.019 \pm .000$ &             $.006 \pm .000$ &    $.008 \pm .000$ &    $.007 \pm .000$ &  $ { .000 \pm .000 }$ &    $.005 \pm .000$ &          \\
KEGGU & $40.7$K & $26$ &          &  $.023 \pm .000$ &             $.013 \pm .000$ &    $.015 \pm .000$ &    $.014 \pm .000$ &  $ { .000 \pm .000 }$ &    $.013 \pm .000$ &          \\
\hline
Precipitation & $64.8$K & $3$ &          &              --- &                         --- &    $.182 \pm .001$ &    $.030 \pm .000$ &  $ { .000 \pm .000 }$ &    $.005 \pm .000$ &          \\
UKHousing & $116$K & $2$ &          &  $.280 \pm .001$ &             $.112 \pm .000$ &    $.345 \pm .002$ &    $.144 \pm .000$ &  $ { .015 \pm .000 }$ &    $.070 \pm .003$ &          \\
3DRoad & $278$K & $3$ &          &  $.584 \pm .000$ &             $.001 \pm .000$ &    $.126 \pm .000$ &    $.203 \pm .001$ &           $ { .000  \pm .000 }$ &    $.001 \pm .000$ &          \\
Covtype & $372$K & $54$ &          &              --- &                         --- &  $ { .000   \pm .000}$ &  $ { .000 \pm .000 }$ &           $ { .000 \pm .000 }$ &  $ { .000  \pm .000}$ &          \\
\bottomrule
\end{tabular}
}
\caption{The learned likelihood noise (mean $\pm$ 1 standard error over 3 random seeds) on high-dimensional datasets (top 8 ones) and spatial datasets (bottom 3 ones). }
\label{tab:uci-prediction-noise}
\end{table*}


\end{document}